\documentclass[10pt,twocolumn,letterpaper]{article}

\usepackage{cvpr}              

\usepackage{booktabs}
\usepackage{multirow}
\usepackage{makecell}  
\usepackage{threeparttable}
\usepackage{graphicx}  
\usepackage{amssymb} 
\usepackage{graphicx} 
\usepackage{footmisc}
\definecolor{LightBlue}{rgb}{0.4, 0.6, 0.9} 
\definecolor{DarkYellow}{rgb}{0.85, 0.65, 0.0}

\PassOptionsToPackage{dvipsnames,table}{xcolor}
\usepackage{textcomp}
\newcommand{\midtilde}{\raisebox{0.5ex}{\texttildelow}}

\definecolor{cvprblue}{rgb}{0.21,0.49,0.74}
\usepackage[pagebackref,breaklinks,colorlinks,allcolors=cvprblue]{hyperref}

\def\confName{CVPR}
\def\confYear{2026}

\title{
Semantic Context Matters: Improving Conditioning for Autoregressive Models
\vspace{-0.85em}
}

\author{
Dongyang Jin$^{*}$,\space\space
Ryan Xu\footnotemark[1] \space , \space
Jianhao Zeng, \space
Rui Lan, \space  \\
Yancheng Bai\footnotemark[3]\hspace{0.4em}\footnotemark[2]\space,  
Lei Sun \hspace{-0.2em}\footnotemark[2]\space, 
and Xiangxiang Chu \\
{\normalsize Amap, Alibaba Group}\\
{\tt \small 12332451@mail.sustech.edu.cn}, 
{\tt \small ryansxu.00@gmail.com}, 
{\tt \small 18826077660@163.com}, 
\\
{\tt \small  \{zengjianhao.zjh, yancheng.byc, ally.sl, chuxiangxiang.cxx\}@alibaba-inc.com}
\vspace{-0.75em}
}

\begin{document}
\maketitle
\footnotetext[1]{Equal contribution.}
\footnotetext[2]{Corresponding author.}
\footnotetext[3]{Project Leader.}

\vspace{-0.5em}
\begin{abstract}
Recently, autoregressive (AR) models have shown strong potential in image generation, offering better scalability and easier integration with unified multi-modal systems compared to diffusion-based methods.
However, extending AR models to general image editing remains challenging due to weak and inefficient conditioning, often leading to poor instruction adherence and visual artifacts.
To address this, we propose \textbf{SCAR}, a \textbf{S}emantic-\textbf{C}ontext-driven method for \textbf{A}utoreg\textbf{R}essive models.
SCAR introduces two key components: \textbf{Compressed Semantic Prefilling}, which encodes high-level semantics into a compact and efficient prefix, and \textbf{Semantic Alignment Guidance}, 
which aligns the last visual hidden states with target semantics during autoregressive decoding to enhance instruction fidelity.
Unlike decoding-stage injection methods, SCAR builds upon the flexibility and generality of vector-quantized-based prefilling while overcoming its semantic limitations and high cost. 
It generalizes across both next-token and next-set AR paradigms with minimal architectural changes.
SCAR achieves superior visual fidelity and semantic alignment on both instruction editing and controllable generation benchmarks, outperforming prior AR-based methods while maintaining controllability.
Code will be released at \url{https://github.com/AMAP-ML/SCAR}.
\end{abstract}

\vspace{-0.5em}
\section{Introduction}
\label{sec:intro}

The advent of large-scale generative models~\cite{sdxl,lan2025flux,he2025ragsr,han2025infinity,liu2025infinitystar} has revolutionized the field of image editing~\cite{sdedit,controlnet}, opening up unprecedented possibilities for creative expression and content manipulation. Within this rapidly evolving landscape, two dominant generative paradigms have emerged: diffusion~\cite{ddpm,dit} and autoregressive (AR) models~\cite{llamagen,var}. Diffusion models, renowned for their powerful text-to-image synthesis capabilities, demonstrate a strong grasp of semantic concepts, allowing for flexible and diverse edits based on user prompts. Concurrently, autoregressive models, owing to their excellent scaling properties and immense potential within unified multimodal generation and understanding model (UMM) architectures~\cite{lumina-mgpt,zhuang2025vargpt}, have achieved generation quality approaching that of top-tier diffusion models, establishing them as a critical and highly promising direction for future research.

\begin{figure}[!t]
\centering
\includegraphics[width=1.0\columnwidth]{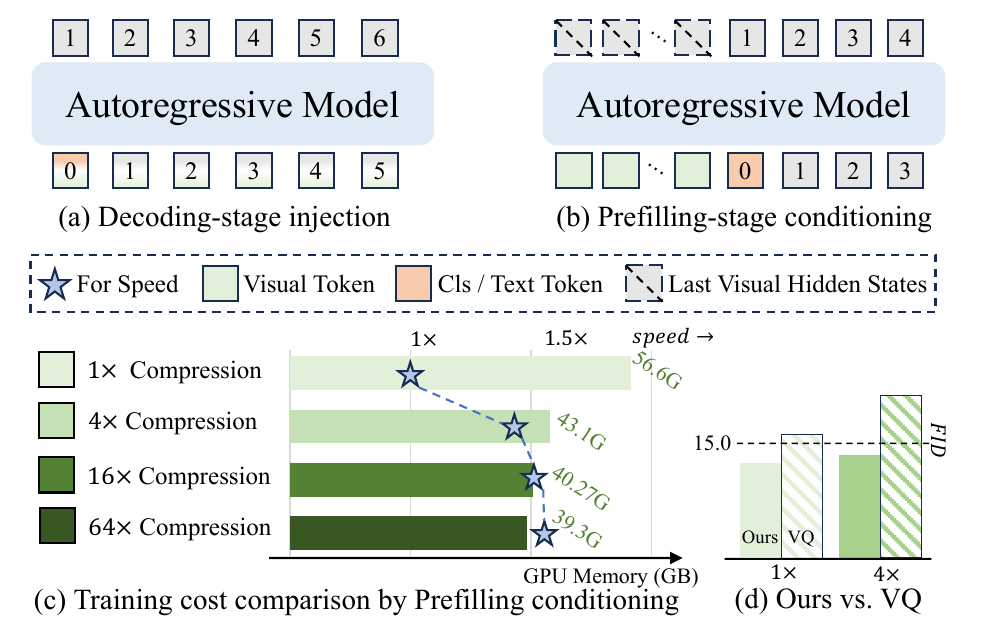}
\vspace{-1.9em}
\caption{
(a) Decoding-stage injection and (b) Prefilling-stage conditioning for condition injection. 
(c) Training cost under different visual token compression. 
4× reduces GPU memory usage by 23.9\% (from 56.6 to 43.1GB) and accelerates training by 1.42×.
(d) Comparison between ours and VQ token prefilling.
}
\vspace{-0.8em}
\label{fig:outline}
\end{figure}

However, translating this raw generative power into versatile, general image editing capabilities remains a significant challenge for AR models. Compared to diffusion-based editing methods~\cite{instructpix2pix,controlnet,controlnet++,unicontrol}, recent AR-based approaches~\cite {li2024controlar,mu2025editar} have yet to achieve the same level of semantic control and general editing scope.
These works can be broadly categorized into two types based on their conditioning strategies: 
(1) \textbf{Decoding-stage injection} (\Cref{fig:outline}.a), as in \texttt{ControlAR}~\cite{li2024controlar}, injects condition signals into the model's intermediate layers. It performs excellent controllable generation, but the injected strong spatial guidance disrupts the autoregressive process for general editing  (see \Cref{fig:t2i_edit}).
(2) \textbf{Prefilling-stage conditioning} (\Cref{fig:outline}.b), employed by works like \texttt{EditAR}~\cite{mu2025editar} and most UMMs~ \cite{xin2025lumina,zhuang2025vargpt,li2025onecat,cao2025hunyuanimage}, prepends visual tokens from the condition image to the input sequence.
However, this method introduces debilitating computational overhead: prepending vector-quantized (VQ) tokens of the condition image can double the sequence length, increasing the computational cost, as shown in~\Cref{fig:outline}.c. Moreover, VQ tokens are widely recognized as sparse-semantic~\cite{harmon,blip3,unitok}, lacking the high-level representation needed for complex edits.

In this work, we revisit the prefilling conditioning paradigm from a semantic perspective and argue that it offers a powerful route toward general image editing and is also compatible with various AR paradigms. 
We identify the prefilling bottleneck as their reliance on inefficient and VQ-based prefixes with shallow semantics. 
Our approach introduces two key innovations. 

First, we propose \textbf{Compressed Semantic Prefilling}, which replaces the inefficient VQ token prefix with compact, rich-semantic vision features extracted from the source image via a frozen vision foundation model (VFM)~\cite{dinov2,clip,vit}. 
We introduce a learnable semantic compression module that drastically reduces the prefilling sequence length (e.g., $4\times$ compression from 1024 to 256 tokens) while preserving essential high-level semantics for editing. 
This significantly enhances computational efficiency and provides the model with a compact semantic understanding of the source content before generation, as shown in~\Cref{fig:outline}.d. 
Notably, our semantic prefix remains robust even when compressed, while VQ token prefix suffers sharp performance drops under similar settings.

Second, to bridge the gap between sparse textual instructions and the desired dense visual guidance, we introduce a novel \textbf{Semantic Alignment Guidance}. 
Instead of relying solely on the sparse text condition, we use the visual features of the target edited image from the VFM as a dense semantic guide. 
Unlike the prior method~\cite{mu2025editar} that distills supervision onto output VQ tokens, we apply an auxiliary constraint objective that aligns the autoregressive model's last visual hidden states with the target's semantic representation, providing a formulation more consistent with the causal decoding process.
This provides a dense, in-context learning signal that steers the model's internal reasoning toward the edit target.

Our contributions can be summarized as follows:
\begin{itemize}[leftmargin=*, itemsep=2pt]
    \item 
    We present a \textbf{S}emantic-\textbf{C}ontext-driven  \textbf{A}utoreg\textbf{R}essive method (\textbf{SCAR}) for general image editing, designed to integrate with most autoregressive models seamlessly.
    
    \item We design Compressed Semantic Prefilling, an efficient visual token compression mechanism to produce compact semantic features for conditioning autoregressive models, overcoming the limitations of VQ-based prefilling.
    
    \item We introduce Semantic Alignment Guidance, providing dense, progressive, in-context semantic guidance to the autoregressive process, enabling complex and semantically-accurate edits.
    
    \item We demonstrate state-of-the-art performance on challenging instruction editing and controllable generation benchmarks, proving SCAR is more effective, efficient, and semantically-aware than existing AR editing frameworks.
\end{itemize}

\section{Related Work}
\subsection{Image Generation}
Two dominant paradigms have emerged in image generation: diffusion models and AR models. 
Diffusion models synthesize images by iteratively denoising Gaussian noise, with DDPM~\cite{ho2020denoising} marking a major breakthrough.
Subsequent work improves generation quality and efficiency~\cite{nichol2021improved,rombach2022high,Wang_2024,zhang2025robust,wang2025jasmine,qu2026scale} through advances in sampling strategies and latent-space modeling.
They have become the backbone of text-to-image and text-to-video generation~\cite{saharia2022photorealistic, singer2022make,ma2024passersby,lan2025flux,zhang2025boow,wang2025editor,zeng2025eevee,chen2026layer}, typically using U-Net for denoising and CLIP~\cite{clip,wei2025hq} or T5~\cite{t5} for text conditioning via cross-attention.
Recent models like DiT~\cite{dit} replace U-Net with Transformer backbones, achieving strong performance.
Despite their success, diffusion models remain computationally expensive, motivating exploration of more efficient alternatives such as autoregressive approaches.

\begin{figure*}[!t]
\centering
\includegraphics[width=2.0\columnwidth]{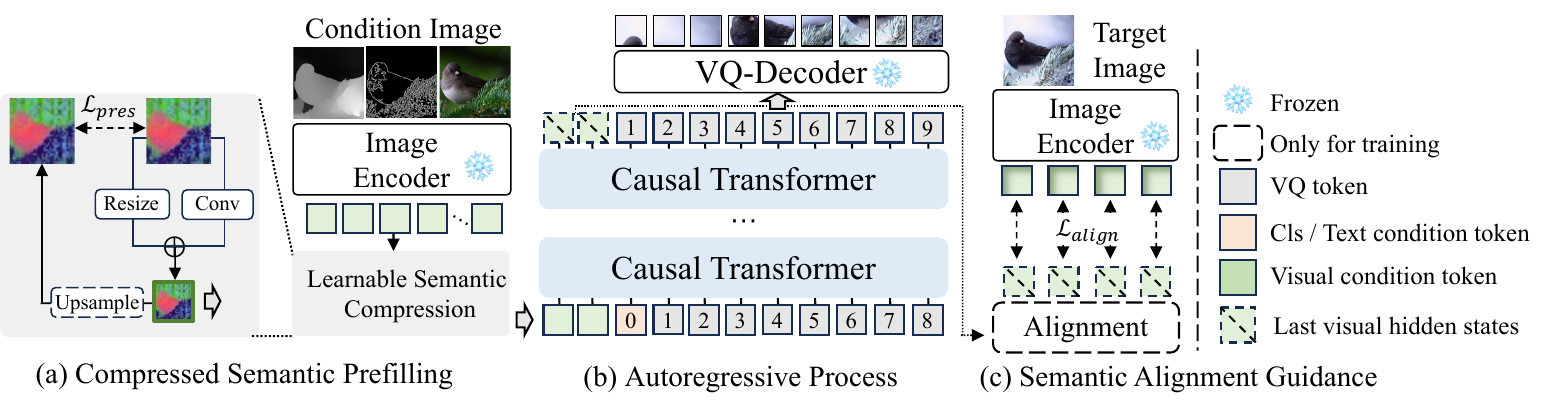}
\vspace{-1em}
\caption{
Overview of our proposed \textbf{SCAR}, a prefilling-based method for autoregressive image editing. 
SCAR is composed of (a) \textbf{Compressed Semantic Prefilling} (see~\Cref{sec:scp} for details) and (c) \textbf{Semantic Alignment Guidance} (see~\Cref{sec:sra} for details), jointly enabling semantically guided generation. The framework is general and compatible with both \textit{next-token} and \textit{next-set} AR paradigms.
}
\label{fig:pipeline}
\vspace{-0.9em}
\end{figure*}

In contrast, autoregressive (AR) models formulate image synthesis as a sequence modeling problem, predicting visual tokens step by step.
Early works~\cite{van2016pixel} focused on pixel-level generation.
Building on the success of large language models~\cite{touvron2023llama, achiam2023gpt}, recent approaches adopt discrete quantizers such as VQ-VAE~\cite{van2017neural} and VQ-GAN~\cite{esser2021taming} to convert image patches into token indices, enabling next-token prediction over visual sequences.
Recent advances have expanded this paradigm in two directions:
(1) conventional next-token AR models~\cite{open-magvit2,ma2025towards,ma2025stage}, such as LlamaGen~\cite{llamagen} and AiM~\cite{aim}, leverage large-scale transformer backbones for high-quality image synthesis; and
(2) Visual Autoregressive Modeling (VAR), which reformulates generation as next-set prediction rather than raster-scan token prediction, offering improved scalability and global coherence~\cite{var, ma2024star, tang2024hart, han2025infinity, zhuang2025vargpt}.
Overall, AR-based approaches have achieved image quality comparable to diffusion models while being more efficient in sampling and deployment.

\subsection{General Image Editing}
General Image editing, including controllable and instruction editing, is a key focus in generative modeling, guided by external signals (e.g., text, edge maps, reference images).
While early approaches incorporated class labels or attributes in GANs~\cite{mirza2014conditional,song2023fashion} and VAEs~\cite{kingma2013auto,zeng2024cat}, recent diffusion-based methods~\cite{controlnet,t2i-adapter,unicontrol} achieve fine-grained control via cross-attention or adapter modules.
Building on this, text-guided instruction editing methods~\cite{prompt2prompt,sdedit,instructpix2pix} further extend controllable generation to image manipulation. However, these approaches often involve costly inversion, require modality-specific tuning, or suffer from limited generalization across tasks and conditions.

In contrast, image editing with autoregressive (AR) models, including next-token and next-set paradigms, remains relatively underexplored.
Recent AR-based methods can be broadly categorized by the stage at which control signals are injected:
\textbf{Decoding-stage injection}~\cite{yao2024car,li2024controlar,xu2025scalar} introduces control signals dynamically during generation, typically via cross-attention or feature modulation. 
These approaches offer strong controllability for pixel-level generation, but their complex architectural modifications often limit generalization to instruction-driven editing.
\textbf{Prefilling-stage conditioning}~\cite{lumina-mgpt,xin2025lumina,zhuang2025vargpt,mu2025editar,qu2025varsr,controlvar} prepends visual tokens from the source image to the input sequence, enabling conditioning before generation. 
This paradigm is adopted by EditAR~\cite{mu2025editar} and VARSR~\cite{qu2025varsr}. 
While simple and model-agnostic, such strategies significantly increase sequence length and attention cost, limiting the model’s capacity for precise and instruction-aware editing. 
Furthermore, directly concatenating raw control tokens (e.g., in ControlVAR~\cite{controlvar}) may disrupt the pretrained generation behavior of the AR backbone. 
The reliance on semantically sparse VQ tokens further limits the effectiveness of prefilling-based approaches~\cite{harmon,blip3,unitok}.
In this paper, we adopt prefilling-stage conditioning and propose a compact semantic method for better guidance. 
Our method applies to both next-token and next-set AR models with minimal architectural changes.

\section{SCAR}
\label{sec:sc3}
In this section, we present SCAR, an efficient conditioning strategy for general AR models, as shown in \Cref{fig:pipeline}. 
We first revisit the image generation with general AR (\Cref{sec:preliminary}). 
Next, we introduce Compressed Semantic Prefilling (\Cref{sec:scp}) for the vision prefix. 
Finally, we improve sparse instruction guidance via Semantic Alignment Guidance in~\Cref{sec:sra}.

\subsection{Preliminary}
\label{sec:preliminary}
Autoregressive (AR) models extend the success of large language models~\cite{touvron2023llama,achiam2023gpt} to image generation and have become a popular approach in vision.
These models convert images into discrete tokens using vector quantization (e.g., VQVAE~\cite{van2017neural}) and generate outputs by predicting tokens conditioned on preceding context.
Recent AR approaches can be broadly categorized into two types: \textit{next-token prediction}, which autoregressively generates individual tokens along a raster-scan sequence~\cite{llamagen,aim}, and \textit{next-set prediction}, which generates a set of tokens following specialized principles (e.g., VAR~\cite{var} predicts next-scale tokens in a coarse-to-fine manner). 
Despite the difference in basic token units (single token vs. a set of tokens), both follow a unified autoregressive formulation:
\begin{equation}
p(\mathbf{z}) = \prod_{i=1}^{N} p(z_i \mid z_{<i}, c),
\end{equation}
where $\mathbf{z} = \{z_1, z_2, \dots, z_N\}$ denotes the sequence of discrete visual token units, and $c$ represents the conditioning input, such as class labels, text prompts, or control features.
The model is trained by minimizing the negative log-likelihood of the token sequence using a cross-entropy loss:
\begin{equation}
\mathcal{L}_{\text{CE}} = \mathbb{E}_{z_i \sim p(z_i)} \left[ - \log p_{\theta}(z_i \mid z_{<i}, c) \right].
\label{eq:ar}
\end{equation}
This general formulation provides a flexible training objective that accommodates a wide range of AR paradigms for generation and editing tasks.

\subsection{Compressed Semantic Prefilling}
\label{sec:scp}
Our primary objective in the prefilling stage is to establish a conditioning prefix that is simultaneously computationally efficient (via short sequence length) and with rich semantics. 
To this end, we leverage powerful pre-trained Vision Foundation Models (VFM) $\mathcal{E}(\cdot)$ to encode the source condition image, and adopt DINOv2~\cite{dinov2} in practice.
However, this choice introduces a direct conflict with our goal of efficiency. 
For a standard $512\times512$ input, they produce a sequence of 1024 tokens. As illustrated in \Cref{fig:outline}, this verbose prefix without compression leads to severe computational and memory overhead. To retain the rich semantics of VFMs while achieving a compact prefix, we propose a novel learnable token compression module.

\noindent \textbf{Learnable Semantic Compression.}
Directly using the semantic feature token $\mathbf{F}_s = \mathcal{E}(\mathbf{I}_s)$ of the source condition image $\mathbf{I}_s$ as a prefix is still inefficient, as $\mathbf{F}_s\in \mathbb{R}^{h \times w \times d}$ can contain over 1024 tokens (e.g., $h\times w = 32 \times 32$), where $d$ is the feature dimension. 
To address this, we introduce a learnable semantic compressive module $\mathcal{P}_k(\cdot)$, which compresses the vision features by a factor of $k$ while preserving high-level semantics. 
Specifically, it compresses the source feature map through a spatial downsampling $\mathcal{R}_k(\cdot)$ and a paralleled stride convolution downsampling $\mathcal{C}_k(\cdot)$:
\begin{equation}
    \mathbf{F}_c = \mathcal{P}_k(\mathbf{F}_s) 
    = \mathcal{C}_k(\mathbf{F}_s)+\mathcal{R}_k(\mathbf{F}_s),
    \label{eq:comp}
\end{equation}
where $\mathbf{F}_c \in \mathbb{R}^{\frac{h}{k} \times \frac{w}{k} \times d}$ 
is the compressed semantic feature.

To ensure $\mathcal{P}_k(\cdot)$ learn to preserve critical semantic information during this compression, we train it with a lightweight upsampling module $\mathcal{U}_k$ implemented as a pixel shuffle operation~\cite{shi2016real,tokenshuffle} with a \textbf{semantic preservation loss}:
\begin{equation}
    \mathcal{L}_{pres} = \| \mathbf{F}_s - \mathcal{U}_k(\mathbf{F}_c) \|_2^2.
    \label{eq:loss_comp}
\end{equation}
This loss, computed in the vision feature space, trains $\mathcal{P}_k(\cdot)$ to retain enough information necessary for semantic reconstruction, discarding redundant details. At inference, $\mathcal{U}_k(\cdot)$ is discarded.

\noindent\textbf{Input Sequence Formulation.}
The final input sequence $\mathbf{S}$ fed to the Causal Transformer $\mathcal{G}_{\theta}$ is the concatenation of the compressed source condition semantic sequence $\mathbf{P}_s$, the target text embedding $\mathbf{T}_t$, and the VQ sequence of the target edited image $\mathbf{Z}_t$:
\begin{equation}
    \mathbf{S} = [\mathbf{P}_s ; \mathbf{T}_t ; \mathbf{Z}_t],
\label{eq:ar}
\end{equation}
where $\mathbf{P}_s \in \mathbb{R}^{L_c \times d}$ is the flattened sequence obtained from the compressed feature map $\mathbf{F}_c$, and $L_c = \tfrac{h \times w}{k^2}$ denotes the reduced token length. We achieve a \textbf{$k^2\times$ compression} compared to directly prefilling.

We also change the attention mask. Specifically, $\mathbf{P}_s$ and $\mathbf{T}_t$ attend to each other bidirectionally, allowing for deep interaction of visual and text semantic information. The quantized tokens $\mathbf{Z}_t$ attend causally to the prefix $[\mathbf{P}_s; \mathbf{T}_t]$ and all preceding VQ tokens, maintaining the autoregressive property for generation.

\subsection{Semantic Alignment Guidance}
\label{sec:sra}
A fundamental challenge in instruction-guided editing is the semantic gap. The text prompt provides only sparse, high-level guidance, which is often insufficient to steer the generation of thousands of dense, low-level VQ tokens $\mathbf{Z}_t$. The standard autoregressive loss $\mathcal{L}_{CE}$ (in~\Cref{eq:ar}) provides supervision on $\mathbf{Z}_t$, but it does not explicitly teach the model how to use the context $[\mathbf{P}_s; \mathbf{T}_t]$ to achieve the desired edit. 
To bridge this gap, we introduce the Semantic Alignment Guidance. 
The core innovation is to provide a dense, in-context learning signal that teaches the model to understand the target semantics $\mathbf{P}_t$ via source semantics $\mathbf{P}_s$ before it even generates the first VQ token.

Specifically, we compute the dense semantic representation of the target image $\mathbf{I}_t$. We use the same frozen, pre-trained $\mathcal{E}(\cdot)$ and our same learnable compressor $\mathcal{P}_k(\cdot)$ to ensure the compressed semantic source and target representations lie in the same space:
\begin{equation}
    \mathbf{P}_t = \mathcal{P}_k\big(\mathcal{E}(\mathbf{I}_t)\big),
\end{equation}
where $\mathbf{P}_t \in \mathbb{R}^{L_c \times d}$ serves as our ground-truth semantic target. In the training phase, we feed the input sequenced $\mathbf{S}$ through the causal transformer $\mathcal{G}_{\theta}$ and extract the last output hidden states $\mathbf{H}_s$ corresponding to the source semantic tokens $\mathbf{P}_s$:
\begin{equation}
    \mathbf{H} = \mathcal{G}_{\theta}(\mathbf{S}), \quad
    \mathbf{H}_s = \mathbf{H}[1:L_c, :],
\end{equation}
where $\mathbf{H}_s$ represents the autoregressive model's internal reasoning of the source image after being modulated by the edit instruction. Our semantic alignment guidance then forces this internal understanding $\textbf{H}_s$ to align with the target semantic representation $\textbf{P}_t$ via a $\ell_2$ constraint:
\begin{equation}
    \mathcal{L}_{align} = \| \mathbf{H}_s - \mathbf{P}_t \|_2^2.
    \label{eq:sra_loss}
\end{equation}
This objective serves as a dense and semantically grounded guidance signal, encouraging the model to align its internal representation of the source condition toward the desired target semantics, providing a rich in-context prior that significantly eases the subsequent prediction of accurate VQ tokens $\mathbf{Z}_t$. Besides, this alignment operates within a unified feature space, ensuring semantic consistency and stable optimization across both source and target representations.

\begin{table*}[!t]
\centering
\caption{
Quantitative results of C2I controllable generation on ImageNet~\cite{deng2009imagenet}. 
Values marked with \midtilde\ are estimated from histograms in the corresponding paper.
Cells highlighted with \colorbox[HTML]{E6F2FF}{\phantom{xx}} and \colorbox[HTML]{D9FFD9}{\phantom{xx}} denote the best performance under the \textbf{next-set} and \textbf{next-token} autoregressive settings, respectively.
\textbf{SCAR-Uni} denotes a unified model trained jointly across all conditions, rather than SCAR for each condition.
}
\vspace{-0.6em}
\resizebox{1.85\columnwidth}{!}{ 
\setlength{\tabcolsep}{1mm}
\begin{tabular}{c|c|c|cc|cc|cc|cc|cc}
\toprule
\multirow{2}{*}{Type}  & \multirow{2}{*}{Method}                         & \multirow{2}{*}{Model} & \multicolumn{2}{c|}{Canny} & \multicolumn{2}{c|}{Depth} & \multicolumn{2}{c|}{Normal} & \multicolumn{2}{c|}{HED} & \multicolumn{2}{c}{Sketch} \\
                       &                                                 &                        & FID↓  & F1-Score↑  & FID↓  & RMSE↓      & FID↓  & RMSE↓       & FID↓ & SSIM↑    & FID↓  & F1-Score↑ \\
\midrule

\multirow{2}{*}{\rotatebox{90}{Diff.}} & T2IAdapter~\cite{mou2024t2i}                    & SD1.5            & \midtilde 10.2 & -          & \midtilde 9.9  & -          & \midtilde 9.5  & -           & \midtilde 9.3  & -        & \midtilde 16.2 & - \\
                       & ControlNet~\cite{zhang2023adding}               & SD1.5                    & \midtilde 11.6 & -          & \midtilde 9.2  & -          & \midtilde 8.9  & -           & \midtilde 8.6  & -        & \midtilde 15.3 & - \\
\midrule

\multirow{18}{*}{\rotatebox{90}{AR}}    & \multirow{3}{*}{ControlAR~\cite{li2024controlar}} & AiM-L                & 9.66  & 30.36      & 7.39  & 35.01      & \multicolumn{2}{c|}{-}          & \multicolumn{2}{c|}{-}        & \multicolumn{2}{c}{-}  \\
                       &                                                 & LlamaGen-B             & 10.64 & 34.15      & 6.67  & 32.41    & \multicolumn{2}{c|}{-}           & \multicolumn{2}{c|}{-}         & \multicolumn{2}{c}{-}   \\
                       &                                                 & \cellcolor[rgb]{0.85,1,0.85}LlamaGen-L             & \cellcolor[rgb]{0.85,1,0.85}7.69  & \cellcolor[rgb]{0.85,1,0.85}34.91      & \cellcolor[rgb]{0.85,1,0.85}4.19  & \cellcolor[rgb]{0.85,1,0.85}31.11 & \multicolumn{2}{c|}{\cellcolor[rgb]{0.85,1,0.85}-}            & \multicolumn{2}{c|}{\cellcolor[rgb]{0.85,1,0.85}-}         & \multicolumn{2}{c}{\cellcolor[rgb]{0.85,1,0.85}-}   \\
                       \cline{2-13}

                       & \multirow{3}{*}{ControlVAR~\cite{controlvar}}   & VAR-d16                & \midtilde 16.2 & -          & \midtilde 13.8 & -          & \midtilde 14.2 & -           & \multicolumn{2}{c|}{-}  & \multicolumn{2}{c}{-}  \\
                       &                                                 & VAR-d20                & \midtilde 13.0 & -          & \midtilde 13.4 & -          & \midtilde 12.8 & -           & \multicolumn{2}{c|}{-}  & \multicolumn{2}{c}{-}  \\
                       &                                                 & \cellcolor[rgb]{0.9,0.95,1}VAR-d30                & \cellcolor[rgb]{0.9,0.95,1}7.85  & \cellcolor[rgb]{0.9,0.95,1}-          & \cellcolor[rgb]{0.9,0.95,1}6.50  & \cellcolor[rgb]{0.9,0.95,1}-          & \cellcolor[rgb]{0.9,0.95,1}6.20  & \cellcolor[rgb]{0.9,0.95,1}-           & \multicolumn{2}{c|}{\cellcolor[rgb]{0.9,0.95,1}-}         & \multicolumn{2}{c}{\cellcolor[rgb]{0.9,0.95,1}-} \\
                       \cline{2-13}

                       & \multirow{3}{*}{CAR~\cite{yao2024car}}          & VAR-d16                & \midtilde 12.8 & -          & \midtilde 10.8 & -          & \midtilde 11.0 & -           & \midtilde 9.8  & -        & \midtilde 13.2 & - \\
                       &                                                 & VAR-d20                & \midtilde 10.2 & -          & \midtilde 8.0  & -          & \midtilde 8.8  & -           & \midtilde 7.2  & -        & \midtilde 11.2 & - \\
                       &                                                 & \cellcolor[rgb]{0.9,0.95,1}VAR-d30                & \cellcolor[rgb]{0.9,0.95,1}8.30  & \cellcolor[rgb]{0.9,0.95,1}-          & \cellcolor[rgb]{0.9,0.95,1}6.90  & \cellcolor[rgb]{0.9,0.95,1}-          & \cellcolor[rgb]{0.9,0.95,1}6.60  & \cellcolor[rgb]{0.9,0.95,1}-           & \cellcolor[rgb]{0.9,0.95,1}5.60  & \cellcolor[rgb]{0.9,0.95,1}-        & \cellcolor[rgb]{0.9,0.95,1}10.20 & \cellcolor[rgb]{0.9,0.95,1}- \\
                       \cline{2-13}
                       & \multirow{4}{*}{\shortstack{SCAR-Uni\\(Ours)}}       & VAR-d16                & 2.77  & 29.41  & 4.29  & 35.61 & 4.14  & 28.51 & 2.40  & 75.29 & 4.59  & 76.88    \\
                       &                                                 & \cellcolor[rgb]{0.9,0.95,1}VAR-d20 & \cellcolor[rgb]{0.9,0.95,1} 2.22 & \cellcolor[rgb]{0.9,0.95,1}29.61  & \cellcolor[rgb]{0.9,0.95,1}3.27& \cellcolor[rgb]{0.9,0.95,1}35.08 & \cellcolor[rgb]{0.9,0.95,1}3.16 & \cellcolor[rgb]{0.9,0.95,1}27.73 & \cellcolor[rgb]{0.9,0.95,1}1.98  & \cellcolor[rgb]{0.9,0.95,1}75.67 &\cellcolor[rgb]{0.9,0.95,1}3.37 &\cellcolor[rgb]{0.9,0.95,1} 77.06    \\  
                       \cline{3-13}
                       &                                                 & LlamaGen-B             & 4.92  & 30.69 & 5.04 & 36.60  & 4.82  & 28.48  & 4.18 & 75.24  & 5.86  & 76.41
                       \\
                       
                       &       & \cellcolor[rgb]{0.85,1,0.85}LlamaGen-L    & \cellcolor[rgb]{0.85,1,0.85}2.54 & \cellcolor[rgb]{0.85,1,0.85}31.27 & \cellcolor[rgb]{0.85,1,0.85}2.77  & \cellcolor[rgb]{0.85,1,0.85}34.35 & \cellcolor[rgb]{0.85,1,0.85} 2.51  & \cellcolor[rgb]{0.85,1,0.85}27.73& \cellcolor[rgb]{0.85,1,0.85}2.39  & \cellcolor[rgb]{0.85,1,0.85} 77.08& \cellcolor[rgb]{0.85,1,0.85}3.02  & \cellcolor[rgb]{0.85,1,0.85}77.51 \\ 
                       \cline{2-13}

                       & \multirow{5}{*}{\shortstack{SCAR\\(Ours)}}       & VAR-d12                & 2.88  & 29.82     & 3.46  & 34.61 & 3.60  & 27.74    & 2.11  & 77.81    & 4.40  & 77.17    \\
                       &                                                   & VAR-d16                & 2.10  & 30.95    & 3.54  &  33.33      & 3.18  & 26.88 & 1.81  & 78.61    &3.83   & 77.83  \\  
                       &                                                 & \cellcolor[rgb]{0.9,0.95,1}VAR-d20 & \cellcolor[rgb]{0.9,0.95,1} 1.97  & \cellcolor[rgb]{0.9,0.95,1}31.27  & \cellcolor[rgb]{0.9,0.95,1}3.29 & \cellcolor[rgb]{0.9,0.95,1}33.32  & \cellcolor[rgb]{0.9,0.95,1}2.96  & \cellcolor[rgb]{0.9,0.95,1}26.73 & \cellcolor[rgb]{0.9,0.95,1}1.51  & \cellcolor[rgb]{0.9,0.95,1}79.72 &\cellcolor[rgb]{0.9,0.95,1}3.39 &\cellcolor[rgb]{0.9,0.95,1}77.74     \\   
                       \cline{3-13}
                       &                                                 & LlamaGen-B             & 5.32  & 31.57     & 4.20  & 34.36  & 4.06  & 26.77 & 3.89  & 78.28 & 4.58 & 77.57
                       \\
                       &                                                 & \cellcolor[rgb]{0.85,1,0.85}LlamaGen-L             & \cellcolor[rgb]{0.85,1,0.85}2.69  & \cellcolor[rgb]{0.85,1,0.85}31.82     & \cellcolor[rgb]{0.85,1,0.85}2.69  & \cellcolor[rgb]{0.85,1,0.85} 32.82 & \cellcolor[rgb]{0.85,1,0.85}2.50  & \cellcolor[rgb]{0.85,1,0.85}26.05      & \cellcolor[rgb]{0.85,1,0.85}2.67  & \cellcolor[rgb]{0.85,1,0.85}79.29 & \cellcolor[rgb]{0.85,1,0.85}3.04  & \cellcolor[rgb]{0.85,1,0.85} 78.30   \\
\bottomrule
\end{tabular}
}
\vspace{-0.1em}
\label{tab:c2i_control}
\end{table*}

\begin{figure*}[!t]
\vspace{-0.8em}
\centering
\includegraphics[width=1.9\columnwidth]{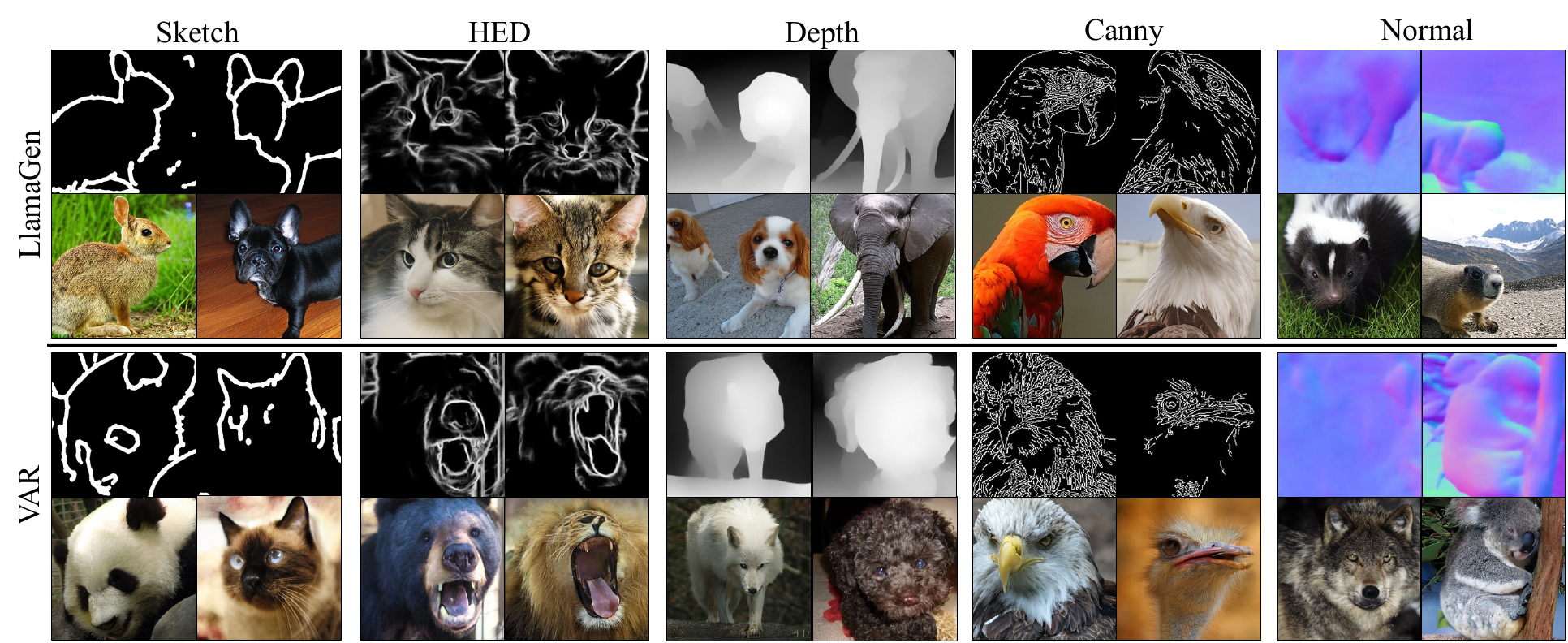}
\vspace{-0.6em}
\caption{
Visualization of C2I controllable generation. 
Our SCAR demonstrates results respectively based on VAR~\cite{var} and LlamaGen~\cite{llamagen}.
}
\vspace{-1em}
\label{fig:c2i_control}
\end{figure*}

\begin{figure}[!t]
\centering
\includegraphics[width=0.9\columnwidth]{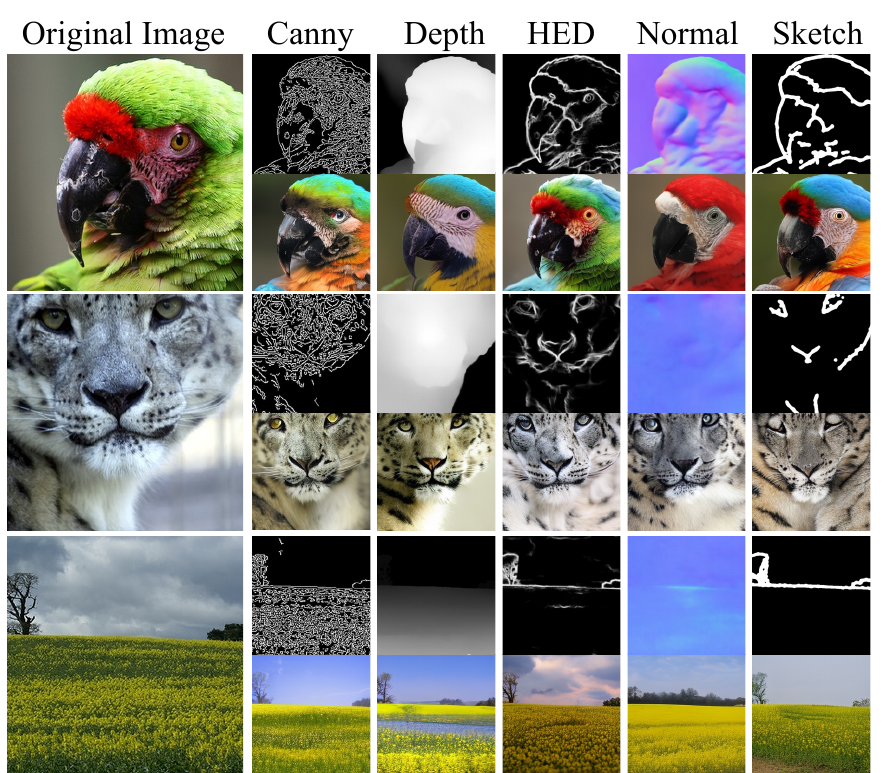}
\vspace{-0.8em}
\caption{
Visualization of multi-condition controllable SCAR-Uni (based on LlamaGen) under varying control conditions.
}
\label{fig:c2i_control_uni}
\vspace{-1.4em}
\end{figure}

\begin{figure}[!t]
\centering
\includegraphics[width=0.9\columnwidth]{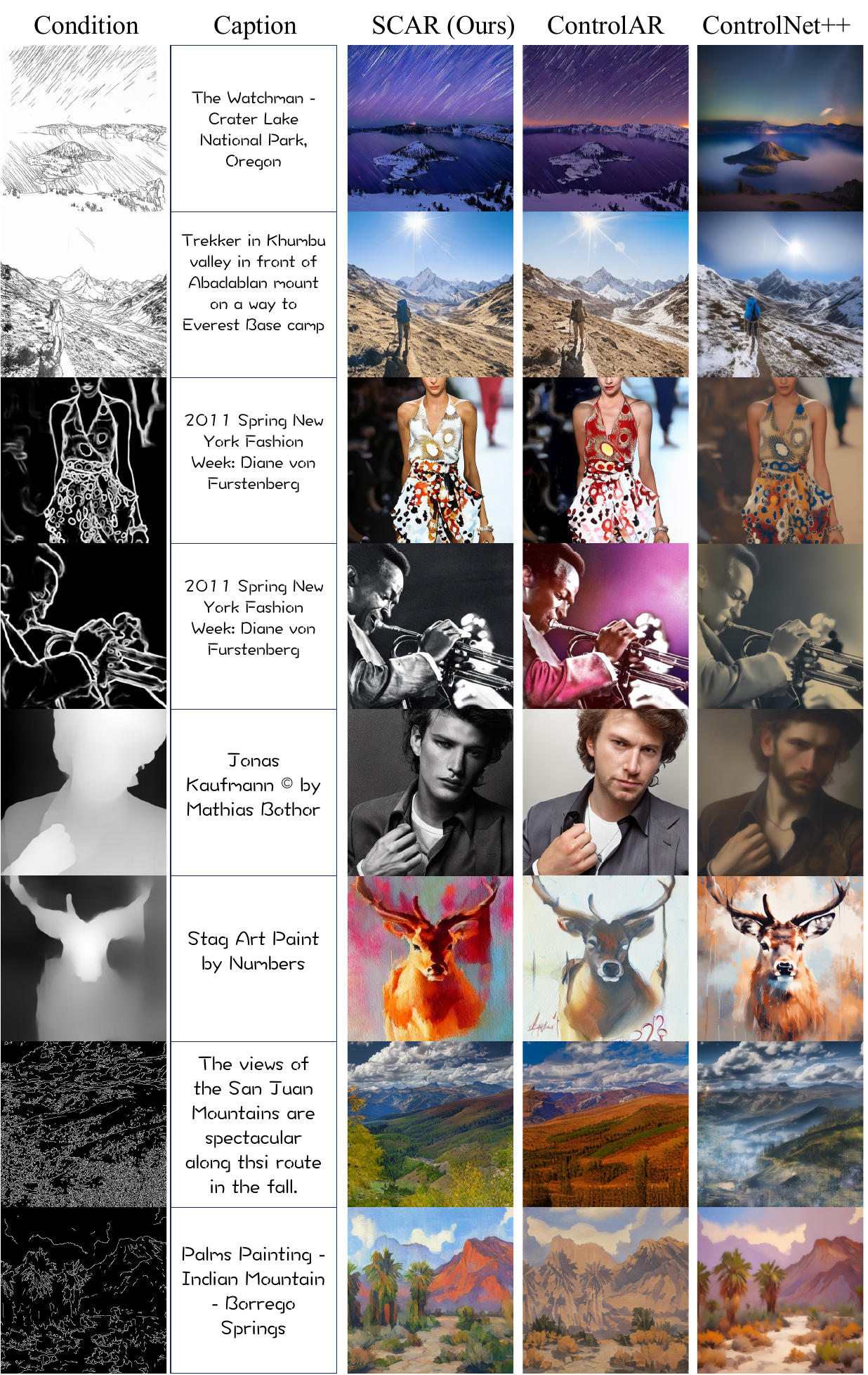}
\vspace{-0.6em}
\caption{
Visualization of T2I controllable generation.
Our SCAR generates images with significantly higher visual quality.
}
\vspace{-1.2em}
\label{fig:t2i_contol}
\end{figure}

\section{Experiments}
\subsection{Experimental Setup}
\noindent \textbf{Dataset.} 
We evaluate our method on controllable generation, covering class-to-image (C2I), text-to-image (T2I), and instruction editing tasks.
For C2I controllable generation, models are trained on ImageNet-256~\cite{deng2009imagenet} with five types of control conditions: Canny, Depth, Normal, HED, and Sketch.
For T2I controllable generation, we use the MultiGen-20M~\cite{unicontrol} for training, considering Depth, Canny, HED, and Lineart as control conditions.
For instruction editing, training is conducted on SEED-Edit-Unsplash~\cite{ge2024seed}.

\noindent \textbf{Training Details.} 
(1) We use DINOv2-B~\cite{dinov2} as the image encoder due to its strong visual representations and scalability.
(2) For C2I controllable generation at a resolution of $256 \times 256$, we adopt two types of models: next-set VAR~\cite{var} and next-token LlamaGen~\cite{llamagen}. 
In this task, SCAR-Uni denotes a variant trained with uniformly sampled control types, supporting any control input at inference.
(3) We use LlamaGen-XL with a T5 encoder~\cite{t5} at $512 \times 512$ for both T2I controllable generation and instruction editing, using captions and editing instructions as text inputs, respectively.
(4) We train VAR for 10 epochs and LlamaGen for 20 epochs on C2I controllable generation, 4 epochs on T2I controllable generation, and 2 epochs on instruction editing.
(5) All models are trained on 8 NVIDIA H20 GPUs.

\begin{table*}[!t]
\centering
\caption{
Image generation quality and conditional consistency of T2I controllable generation on MultiGen-20M~\cite{unicontrol}. 
Our SCAR is based on LlamaGen-XL.
\textbf{Bold} indicates the best result and \underline{underline} indicates the second best. 
The results are conducted on 512 × 512 resolution.
}
\vspace{-0.6em}
 \resizebox{1.8\columnwidth}{!}{ 
\begin{tabular}{c|c|cc|cc|cc|cc}
\toprule
\multirow{2}{*}{Type} & \multirow{2}{*}{Method} & \multicolumn{2}{c|}{Depth}      & \multicolumn{2}{c|}{HED}       & \multicolumn{2}{c|}{Canny}      & \multicolumn{2}{c}{Lineart}    \\
                      &                         & FID↓            & RMSE↓           & FID↓           & SSIM↑           & FID↓            & F1-Score↑       & FID↓           & SSIM↑           \\ \midrule
\multirow{6}{*}{\rotatebox{90}{Diff.}} & GLIGEN~\cite{li2023gligen}                  & 18.36          & 38.83          & \multicolumn{2}{c|}{-}         & 18.89          & 26.94          & \multicolumn{2}{c}{-}          \\
                      & T2I-Adapter~\cite{mou2024t2i}             & 22.52          & 48.40          & \multicolumn{2}{c|}{-}         & 15.96          & 23.65          & \multicolumn{2}{c}{-}          \\
                      & ControlNet~\cite{zhang2023adding}              & 17.76          & 35.90          & 15.41         & 76.21          & 14.73          & 34.65          & 17.44         & 70.54          \\
                      & Uni-ControlNet~\cite{zhao2023uni}          & 20.27          & 40.65          & 17.08         & 69.10          & 17.14          & 27.32          & \multicolumn{2}{c}{-}          \\
                      & UniControl~\cite{qin2023unicontrol}              & 18.66          & 39.18          & 15.99         & 79.69          & 19.94          & 30.82          & \multicolumn{2}{c}{-}          \\
                      & ControlNet++~\cite{li2024controlnet++}            & 16.66          & \textbf{28.32} & 15.01         & 80.97          & 18.23          & \underline{37.04}    & 13.88         & \textbf{83.99} \\ \midrule
\multirow{3}{*}{\rotatebox{90}{AR}}   & ControlAR~\cite{li2024controlar}      & \underline{ 14.61}    & 29.01          & \underline{ 10.53}   & \textbf{85.63} & 17.51          & \textbf{37.08} & \underline{12.41}         & \underline{79.22}    \\
                      & EditAR~\cite{mu2025editar}                  & 15.97          & 34.93          & \multicolumn{2}{c|}{-}         & 13.91          & -              & \multicolumn{2}{c}{-}          \\
                      & SCAR (Ours)                    & \textbf{13.77} & \underline{ 28.89}    & \textbf{8.41} & \underline{ 83.09}    & \textbf{10.82} & 32.46          & \textbf{8.91} & 73.52          \\ \bottomrule
\end{tabular}
}
\vspace{-1.6em}
\label{tab:t2i_control}
\end{table*}

\noindent \textbf{Evaluation and Metrics.}
For controllable generation, we evaluate models based on two main aspects: conditional consistency and image generation quality.
Conditional consistency is measured by comparing the input condition image with the condition image extracted from the generated output.
We use F1-Score for Canny and Sketch, Root Mean Square Error (RMSE) for Normal and Depth, and Structural Similarity Index Measure (SSIM) for HED and Lineart.
Image generation quality is assessed using Fréchet Inception Distance (FID)~\cite{heusel2017gans}.
Class-to-image (C2I) evaluations are conducted on the full $50$K validation set of ImageNet-256, while text-to-image (T2I) evaluations are performed on the official $5$K validation set of MultiGen-20M~\cite{unicontrol}.

For instruction editing, we adopt the PIE-Bench~\cite{ju2023direct}, containing 700 samples covering 10 editing types. 
SCAR takes the source image and editing instructions as input to predict the target image, and follows the evaluation protocol in~\cite{ju2023direct}.
Evaluation metrics cover three aspects: structure consistency, background preservation, and CLIP-based image-text alignment.

\subsection{Controllable Generation Results}
\noindent \textbf{C2I Controllable Generation.}
We evaluate the C2I controllable generation performance of our SCAR and its unified multi-condition variant SCAR-Uni on ImageNet~\cite{deng2009imagenet}.
As shown in~\Cref{tab:c2i_control}, we assess both conditional consistency and image generation quality under two autoregressive paradigms: next-token-based LlamaGen and next-set-based VAR.
Compared with existing C2I controllable generation methods, including diffusion-based methods~\cite{zhang2023adding, mou2024t2i}, next-token AR methods~\cite{li2024controlar}, and next-set AR methods~\cite{controlvar, yao2024car}, SCAR achieves significantly better image quality (e.g., FID \textbf{1.97} vs. 7.69 on Canny), while maintaining competitive control accuracy compared to models that apply decoding-stage injection.
SCAR is validated under both next-token and next-set paradigms, demonstrating the effectiveness and generalizability. 
\Cref{fig:c2i_control} and~\Cref{fig:c2i_control_uni} further illustrate this, showing visual results of SCAR across different autoregressive paradigms, and SCAR-Uni under a unified multi-condition setting.

\begin{table*}[t!]
\centering 
\caption{
Quantitative comparison between our SCAR and other methods on PIE-Bench~\cite{ju2023direct}. 
Metrics include structure consistency, background preservation, and CLIP-based image-text alignment. 
ControlAR* is a retrained variant of ControlAR for editing. 
}
\vspace{-0.6em}
\resizebox{1.85\columnwidth}{!}{ 
\begin{tabular}{c|c|c|ccccccc}
\toprule
\multirow{2}{*}{Type} & \multirow{2}{*}{Method}& \multirow{2}{*}{Model} & Structure & \multicolumn{4}{c}{Background Preservation} & \multicolumn{2}{c}{CLIP Similarity} \\
           \cmidrule(lr){4-4}           \cmidrule(lr){5-8} \cmidrule(lr){9-10}
                      &                        &                         & Distance↓  & PSNR↑    & LPIPS↓      & MSE↓     & SSIM↑   & Whole↑        & Edited↑           \\
\midrule
\multirow{4}{*}{\rotatebox{90}{Diff.}}&InstructPix2Pix~\cite{instructpix2pix} & SD1.5 & 107.43 & 16.69 & 271.33 & 392.22 & 68.39 & 23.49 & 22.20 \\
                                      &InstructDiffusion~\cite{geng2024instructdiffusion} & SD1.5 & 74.21 & 20.88 & 142.35 & 353.45 & 76.70 & 24.06 & 21.57 \\
                                      &MGIE~\cite{fu2023guiding} & SD1.5 & 67.41 & 21.20 & 142.25 & 295.11 & \textbf{77.52} & 24.28 & 21.79 \\
                                      &SEED-X-Edit~\cite{ge2024seed}       & SDXL  & 61.69 & 18.80 & 173.63 & 209.05 & 75.13 & \textbf{25.51} & \underline{21.87} \\
                                      \midrule
\multirow{3}{*}{\rotatebox{90}{AR}}  &ControlAR*~\cite{li2024controlar} & LlamaGen &  116.99 & 14.63 & 289.34 & 590.63 & 63.03 & 23.12 & 21.43  \\ 
                                    &EditAR~\cite{mu2025editar}        & LlamaGen &  \underline{39.43} & \underline{21.32} & \underline{117.15} & \underline{130.27} & 75.13 & 24.87 & \underline{21.87} \\
                                      &SCAR (Ours) & LlamaGen &  \textbf{30.98} & \textbf{22.59} & \textbf{105.09} & \textbf{83.47} & \underline{76.73} & \underline{25.08} & \textbf{21.88}  \\ 
\bottomrule
\end{tabular}%
}
\label{tab:image-editing}
\end{table*}

\begin{figure*}[!t]
\vspace{-0.8em}
\centering
\includegraphics[width=2.0\columnwidth]{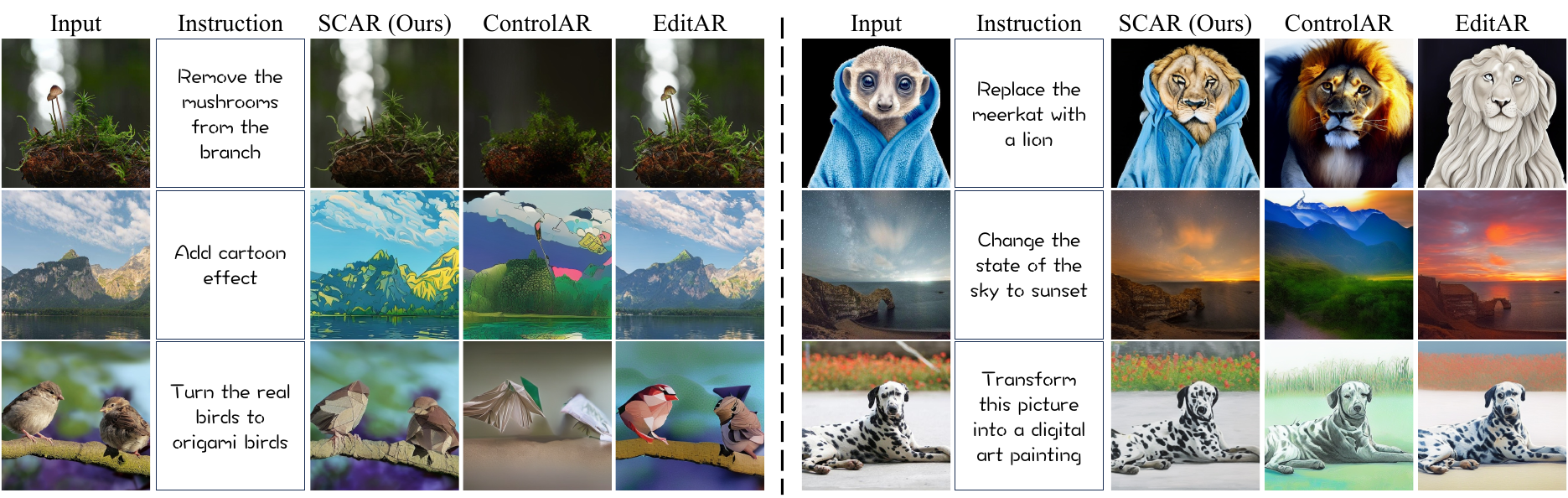}
\vspace{-0.8em}
\caption{
Qualitative results of instruction editing.
Compared with other AR-based methods (ControlAR~\cite{li2024controlar}, EditAR~\cite{mu2025editar}), SCAR can follow instructions more closely and preserve the original image content better.
\textit{Note}: more visualizations can be found in the \textbf{Appendix}.
}
\vspace{-0.8em}
\label{fig:t2i_edit}
\end{figure*}

\noindent \textbf{T2I Controllable Generation.}
We adopt LlamaGen-XL as AR model for SCAR in T2I controllable generation.  
\Cref{tab:t2i_control} reports generation quality (FID) and controllability results across four control conditions on MultiGen-20M~\cite{unicontrol}. 
Alongside representative diffusion-based methods~\cite{li2023gligen,t2i-adapter,zhang2023adding,uni-controlnet,unicontrol, li2024controlnet++}, we compare SCAR with recent AR-based methods, including EditAR~\cite{mu2025editar} and ControlAR~\cite{li2024controlar}.  
SCAR consistently outperforms all competing methods in FID across all conditions, achieving notably lower scores (e.g., HED: \textbf{8.41} vs. 10.53; Depth: \textbf{13.77} vs. 14.61; \textbf{Our SCAR} vs. ControlAR).
While the decoding-stage injection method ControlAR achieves strong alignment, SCAR still delivers highly competitive consistency across HED, Depth, and others.
Notably, under the same backbone, SCAR achieves a better balance between control accuracy and visual fidelity compared to prior methods.
\Cref{fig:t2i_contol} shows qualitative comparisons, where SCAR generates more natural and visually appealing results while maintaining comparable controllability.

\subsection{Instruction Editing Results}
\noindent \textbf{Quantitative Results.} 
\Cref{tab:image-editing} presents quantitative results on PIE-Bench~\cite{ju2023direct}, evaluating structure preservation, background reconstruction, and image-text alignment.  
Under the same backbone architecture (LlamaGen-XL), prefilling-based methods (SCAR and EditAR) substantially outperform decoding-stage injection (ControlAR). 
This indicates decoding-stage injection is suboptimal for instruction editing, as its strong spatial guidance tends to disrupt the autoregressive editing.
SCAR achieves the best results on most metrics, reducing structure distance by 21.4\%, LPIPS by 10.3\%, and MSE by 35.9\%, while improving PSNR by 1.27 dB and boosting regional CLIP similarity.
These gains stem from SCAR's compressed semantic prefilling, which replaces VQ tokens with DINO features that better capture structural and semantic cues.
Furthermore, Semantic Alignment Guidance further enhances instruction-result consistency.

\noindent \textbf{Qualitative Results.} 
As shown in~\Cref{fig:t2i_edit}, we qualitatively compare our SCAR with AR-based methods~\cite{li2024controlar,mu2025editar}. 
SCAR consistently follows instructions while better preserving untouched regions, avoiding unnecessary over-editing. 
Compared to ControlAR and EditAR, it handles fine-grained, structurally complex edits with greater precision, producing clearer boundaries and more natural textures.
These results highlight SCAR's advantages in semantic alignment, structural consistency, and visual quality.
More visualizations can be found in the Appendix

\subsection{Ablation Study}
\noindent \textbf{Compression Strategies.}
We compare different feature compression strategies with a $4\times$ compression (from 1024 to 256 tokens) in~\Cref{tab:ablation_downsample_strategy}.
Experimental results demonstrate that our proposed Learnable Semantic Compression strategy preserves more informative features and fine details during token compression. 
The semantic preservation loss in~\Cref{eq:loss_comp} is introduced during training to guide the compression module to retain essential information for semantic recovery while discarding low-level noise and redundant details, which further enhances generation quality.

\noindent \textbf{Compression Ratio.}
After adopting Learnable Semantic Compression, we further study the impact of different compression ratios.
As shown in~\Cref{tab:ablation_downsample_factor}, $4\times$ achieves comparable quality and consistency to the non-compressed setting ($1\times$), while significantly outperforming $16\times$.
This indicates that $4\times$ compression effectively reduces the input tokens while preserving performance. 
Moreover, as shown in~\Cref{fig:outline}, its generation speed is close to $16\times$, offering a good trade-off between efficiency and quality.
We thus adopt $4\times$ as the default setting.

\begin{table}[!t]
\centering
\caption{
Ablation on different compression strategies. 
All models are trained for 1 epoch (including ~\Cref{tab:ablation_downsample_factor} and~\Cref{tab:ablation_image_encoder}).
}
\vspace{-0.6em}
\resizebox{0.9\columnwidth}{!}{ 
\begin{tabular}{c|c|cc|cc}
 \toprule
\multirow{6}{*}{\rotatebox{90}{MultiGen-20M}} & \multirow{2}{*}{Strategy} & \multicolumn{2}{c|}{HED} & \multicolumn{2}{c}{Depth} \\
                              &                         & FID↓      & SSIM↑         & FID↓         & RMSE↓        \\
                              \cmidrule{2-6}
                              & Resize                  &  10.07    & 80.15         & 15.78       & 34.20      \\
                              & PixelUnshuffle            & \underline{9.82}    & \underline{81.65}         & 15.48       & 34.27            \\
                              & Ours                    & 9.89       & 81.47         & \underline{15.21} &  \textbf{33.90}          \\
                              & Ours + \textit{w/ $\mathcal{L}_{pres}$}            &   \textbf{9.43}          & \textbf{81.76}         & \textbf{14.70}    &  \underline{33.95}          \\
                              
\bottomrule
\end{tabular}
}
\vspace{-0.6em}
\label{tab:ablation_downsample_strategy}
\end{table}

\begin{table}[!t]
\centering
\caption{
Ablation on different compression ratios. 
}
\vspace{-0.6em}
\resizebox{0.95\columnwidth}{!}{ 
\begin{tabular}{c|c|cc|cc}
\toprule
\multirow{5}{*}{\rotatebox{90}{MultiGen-20M}}& \multirow{2}{*}{\begin{tabular}[c]{@{}c@{}}Compression\\Ratio $k^2$ for \cref{eq:comp}\end{tabular}} & \multicolumn{2}{c|}{HED} & \multicolumn{2}{c}{Depth} \\
&                        & FID↓      & SSIM↑      & FID↓        & RMSE↓      \\
\cmidrule{2-6}
&1$\times$              & \textbf{9.29} &\textbf{81.95}&  \textbf{14.61}      &   \underline{34.09}          \\
&4$\times$            &\underline{9.43}&\underline{81.76}  &  \underline{14.70}   &   \textbf{33.95}    \\
&16$\times$              & 10.74     & 79.66           &  16.10      &  34.92           \\ 
\bottomrule
\end{tabular}
}
\vspace{-0.6em}
\label{tab:ablation_downsample_factor}
\end{table}

\begin{table}[!t]
\centering
\caption{
Comparison of Image Encoders $\mathcal{E}$ on MultiGen-20M.
}
\vspace{-0.6em}
\resizebox{1.0\columnwidth}{!}{ 
\begin{tabular}{c|c|c|cc|cc}
\toprule
\multirow{2}{*}{Idx} & \multirow{2}{*}{\begin{tabular}[c]{@{}c@{}}Image\\ Encoder $\mathcal{E}$\end{tabular}} & \multirow{2}{*}{\begin{tabular}[c]{@{}c@{}}Para.\\ of $\mathcal{E}$\end{tabular}} & \multicolumn{2}{c|}{HED} & \multicolumn{2}{c}{Depth} \\
                     &                                                                            &            & FID↓       & SSIM↑       & FID↓         & RMSE↓        \\
\midrule
(a)                  & DINOv2-S                                                                   & 22.1M      & 10.25          &  81.19              &  \underline{15.26}      &  \underline{34.14}      \\
(b)                  & ViT-S                                                                      & 21.8M      & 12.37          & 72.01               &  17.56      &  36.74      \\
(c)                  & ViT-B                                                                      & 86.4M      & 11.87          & 77.45               &  16.86      &  36.39      \\
(d)                  & SAM-B                                                                      & 89.6M      & \underline{10.20}&\underline{81.58}  &  15.57      &  34.94      \\
(e)                  & CLIP-B                                                                     & 149.6M     & 13.78          & 55.43               &  18.35      &  38.10      \\
(f)                  & DINOv2-B                                                                   & 86.6M      & \textbf{9.43}& \textbf{81.76} &  \textbf{14.70}      &  \textbf{33.95}     \\
\bottomrule
\end{tabular}
}
\vspace{-0.8em}
\label{tab:ablation_image_encoder}
\end{table}

\noindent \textbf{Ablations on Image Encoder $\mathcal{E}$.}
In~\Cref{tab:ablation_image_encoder}, we conduct experiments using different image encoders towards different controls on MultiGen-20M~\cite{unicontrol}, including HED and Depth.
Firstly, unlike previous approaches~\cite{li2024controlar,yao2024car}, Our methods adopt frozen image encoders~\cite{vit,dinov2,clip,sam} to preserve the robust features obtained from large-scale pretraining. 
As shown in~\Cref{tab:ablation_image_encoder} (a) to (c), DINOv2~\cite{dinov2} outperforms other pretrained vision models such as ViT~\cite{vit}, CLIP~\cite{clip}, and SAM~\cite{sam}.
Furthermore, comparing (b), (c), (a), and (f) indicates that increasing the scale of the image encoder significantly boosts performance.
Therefore, we choose DINOv2-B as the final image encoder.

\begin{figure}[!t]
\centering
\includegraphics[width=1.0\columnwidth]{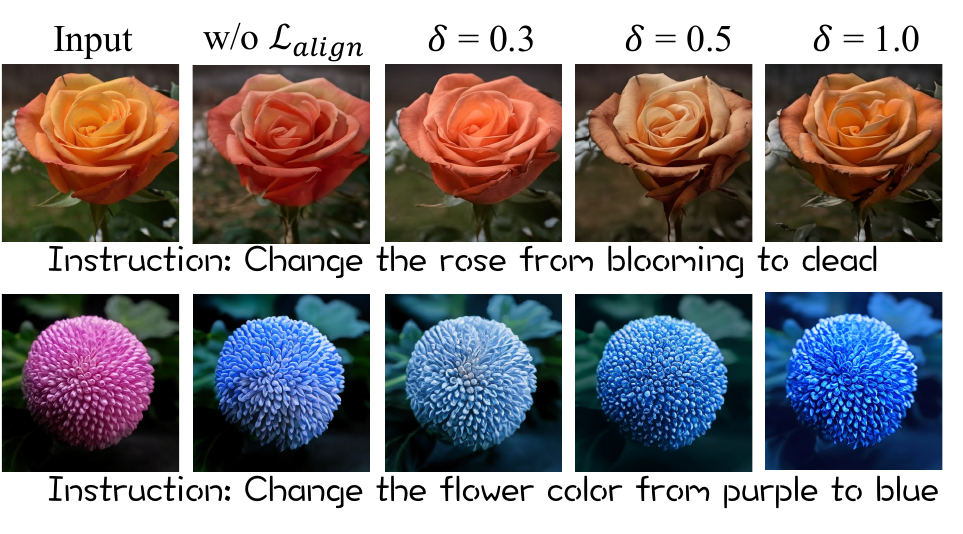}
\vspace{-2.2em}
\caption{
Ablation studies on Semantic Alignment Guidance $\mathcal{L}_{align}$ in~\Cref{eq:sra_loss}.
}
\vspace{-0.8em}
\label{fig:ablation_align}
\end{figure}

\noindent \textbf{Ablations on Semantic Alignment Guidance.}
As shown in~\Cref{fig:ablation_align}, we perform ablation studies on the proposed Semantic Alignment Guidance to assess its effectiveness in improving instruction adherence. 
As shown in~\Cref{fig:ablation_align}, we conduct ablation studies on Semantic Alignment Guidance to assess its effect on instruction fidelity.
Without the alignment loss $\mathcal{L}_{align}$, the model sometimes fails to follow the editing instructions, leading to semantic mismatches. 
Introducing the Semantic Alignment Guidance significantly enhances the model’s ability to follow instructions, with outputs better aligned to the target semantics. 
As the alignment weight $\delta$ increases, instruction fidelity improves; however, overly strong guidance (e.g., $\delta = 1.0$) may cause artifacts such as structural distortion (row 1) or color spillover (row 2). 
We find that setting $\delta = 0.5$ offers the best trade-off between semantic consistency and visual quality, highlighting the effectiveness of our alignment strategy.

\section{Conclusion and Future Work}
\noindent\textbf{Conclusion.} 
In this work, we present \textbf{SCAR}, a prefilling-based conditioning method adaptable to both \textit{next-token} and \textit{next-set} autoregressive models.
To overcome the limitations of existing AR-based editing methods, SCAR introduces two key innovations: 
\textbf{Compressed Semantic Prefilling}, which injects compact and semantically-rich DINO features for improving conditioning; 
and \textbf{Semantic Alignment Guidance}, which provides dense supervision by aligning intermediate hidden states with target image semantics. 
These designs significantly improve both editability and controllability across diverse tasks. 
Our SCAR demonstrates superior performance in controllable generation and instruction editing, while remaining compatible with modern unified AR architectures. 
We hope our work inspires further research into semantic and efficient conditioning strategies for autoregressive models.

\noindent\textbf{Future Work.} 
(1) Scaling SCAR to larger parameter sizes may enhance performance further, following the scaling law of autoregressive models to improve semantic understanding and controllability.
(2) Extending SCAR beyond general image editing to broader generative tasks, such as unified multimodal models and video editing.

{
    \small
    \bibliographystyle{ieeenat_fullname}
    \bibliography{main}
}

\clearpage
\appendix
\section{More Discussion}

\noindent \textbf{Decoding-stage Injection vs. Prefilling-stage condition.}
Conditioning strategies for autoregressive (AR) models fall into two categories: decoding-stage injection and prefilling-stage conditioning.
We argue that the latter provides a more suitable conditioning mechanism for instruction editing and controllable generation, while remaining naturally compatible with both next-token and next-set AR paradigms.

For controllable generation, decoding-stage methods~\cite{li2024controlar,yao2024car,xu2025scalar,liu2025scaleweaver} like ControlAR~\cite{li2024controlar} and CAR~\cite{yao2024car} inject control signals into intermediate layers, enabling fine-grained spatial alignment. However, this tight coupling often leads to training instability and overfitting to local structures, resulting in blurry textures and reduced realism.
In contrast, prefilling-based methods such as EditAR~\cite{mu2025editar} and our SCAR prepend control features as conditioning tokens before decoding. Though offering slightly weaker pixel-level control, this design leads to sharper, more natural outputs and better generalization across control types.
More importantly, for instruction editing, decoding-stage injection fails to align generation with high-level semantics, often producing overconstrained or inconsistent results. It also conflicts with unified multimodal models (UMM)~\cite{an2024mc,luo2024llm,ren2024ultrapixel,peng2024efficient,lin2024draw,lin2025perceiveanythingrecognizeexplain,an2025unictokens,xu2025camel}, as spatial injection disrupts the shared AR generation.
Prefilling-stage conditioning avoids this issue, integrates cleanly with UMM, and better preserves semantic intent during generation.

\noindent \textbf{Semantic Prefix vs. VQ Prefix.}
Table~\ref{tab:vq} reports the quantitative comparison between semantically aligned DINO tokens and standard VQ tokens under two compression settings ($k^2{=}1\times$ and $4\times$) on MultiGen-20M~\cite{unicontrol}.
Across both settings, DINO tokens consistently outperform VQ tokens, underscoring the effectiveness of our semantic prefix for autoregressive editing. 
Without the compression ($1\times$ compression) setting, DINO tokens improve SSIM by +1.96 and reduce FID by 0.91, indicating better perceptual quality and semantic coherence. Under the $4\times$ compression setting, the improvement becomes even more pronounced: SSIM increases by +3.21, and FID drops by 2.83. These results demonstrate that our method not only improves generation quality in standard setups but also maintains strong performance under token compression.

\begin{table}[!t]
\centering
\caption{
Comparison between the semantical DINO token and the VQ token under different compression ratios. 
All models are trained for 1 epoch.
Values in parentheses indicate relative improvement over VQ tokens.
}
\vspace{-0.6em}
\resizebox{0.95\columnwidth}{!}{ 
\begin{tabular}{c|c|cc}
\toprule
\multirow{2}{*}{Prefix Condition} & \multirow{2}{*}{\begin{tabular}[c]{@{}c@{}}Compress\\ Ratio $k^2$\end{tabular}} & \multicolumn{2}{c}{HED} \\
                                  &          & FID↓                            & SSIM↑                             \\ \midrule
VQ token                          & 1$\times$       & 10.20                           & 79.99                              \\
DINO token                        & 1$\times$       & 9.29 \textcolor{BrickRed}{(-0.91)}  & 81.95 \textcolor{ForestGreen}{(+1.96)}  \\ \midrule
VQ token                          & 4$\times$       & 12.26                           & 78.55                              \\
DINO token                        & 4$\times$       & 9.43 \textcolor{BrickRed}{(-2.83)} & 81.76 \textcolor{ForestGreen}{(+3.21)}  \\ \bottomrule
\end{tabular}
}
\vspace{-0.6em}
\label{tab:vq}
\end{table}


\noindent \textbf{Discussion on Semantic Alignment.}
Recent work increasingly suggests that enforcing semantic representation alignment can lead to more effective learning in generative models.
In diffusion models, REPA~\cite{yu2025repa} aligns internal features with pretrained semantic encoders to stabilize training and improve generation. 
REG~\cite{reg} takes a different path by injecting discriminative semantics directly via spatial concatenation with a class token, improving quality and convergence with minimal overhead.
VA-VAE~\cite{vavae} enhances latent interpretability via representation alignment in VAEs. MaskDiT~\cite{zheng2023fast} enforces semantic consistency through masked reconstruction or auxiliary decoders. 
Multi-stage methods~\cite{pernias2023wurstchen,li2024return} leverage pretrained representations as intermediate maps, while USP~\cite{chu2025usp} aligns masked latents in a shared VAE space to unify generation and understanding. 

While these strategies have shown strong results in diffusion-based models, extending semantic alignment to autoregressive (AR) frameworks for editing presents unique challenges. 
EditAR~\cite{mu2025editar} attempts this by distilling supervision from DINO features onto the hidden states of generated VQ tokens. However, as this supervision is applied post-generation, it lacks explicit semantic flow during decoding.

In contrast, SCAR aligns semantic features at the prefilling stage, by matching conditional representations from the source image with a pretrained visual space (e.g., DINO~\cite{dinov2}) before decoding begins. 
This design fits naturally with the AR model's causal structure, introducing dense supervision without interfering with token-level predictions. 
As a result, the model internalizes the semantic correspondence early and propagates it consistently throughout generation. 
Compared to global alignment strategies in diffusion or post-hoc distillation in EditAR, SCAR provides a more direct and effective mechanism for integrating conditional guidance, especially in instruction-based editing where semantic consistency across text, condition, and output is critical.

\begin{figure}[!t]
\centering
\includegraphics[width=1.0\columnwidth]{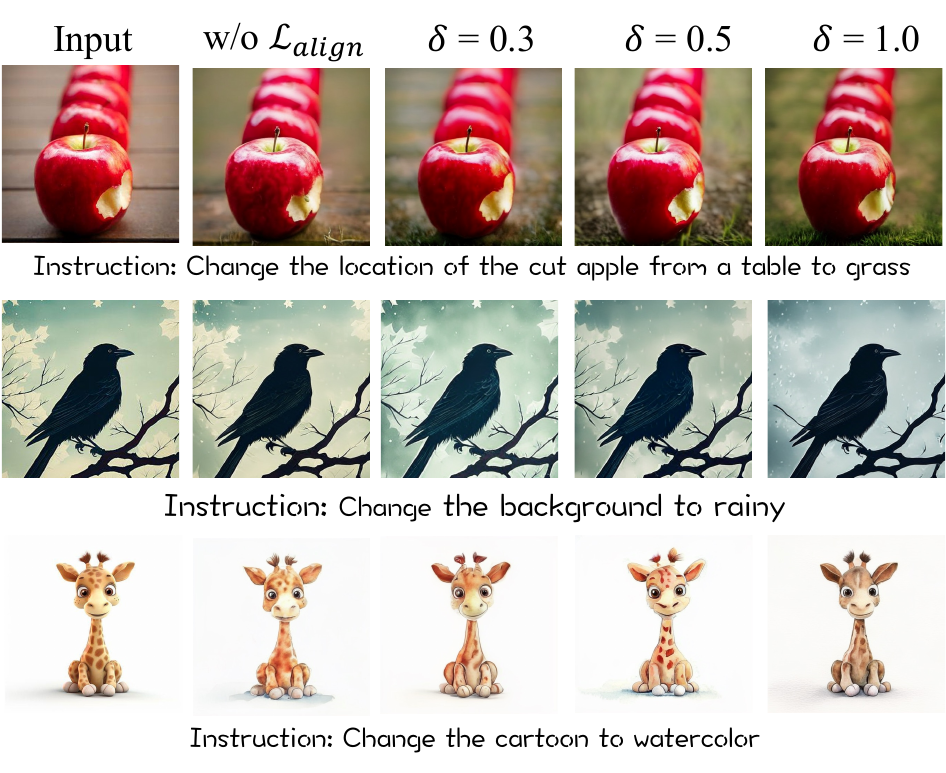}
\vspace{-2.0em}
\caption{
Additional visualizations on Semantic Alignment Guidance $\mathcal{L}_{align}$ in Equation (8).
}
\vspace{-0.6em}
\label{fig:ablation_align_app}
\end{figure}

\section{More Visualization}
\subsection{About Semantic Alignment Guidance}
As shown in~\Cref{fig:ablation_align_app}, without Semantic Alignment Guidance, the model may sometimes produce incomplete or ambiguous edits.
Semantic Alignment Guidance mitigates this by strengthening the correspondence between input conditions and target semantics.
In the apple relocation example (top row), both the \textit{w/o} $\mathcal{L}_{\text{align}}$ and $\delta{=}0.3$ results retain table textures, indicating insufficient background change.
In contrast, $\delta{=}0.5$ successfully introduces grass while preserving the foreground.
However, $\delta{=}1.0$ causes a global green tint, suggesting semantic leakage.
In the background editing task (middle row), $\delta{=}1.0$ enhances the rainy effect but removes fine details such as leaves, while $\delta{=}0.5$ better balances visual accuracy and structure.
For stylization (bottom row), larger $\delta$ values improve the watercolor effect, but $\delta{=}1.0$ leads to facial distortion and oversaturation.

Overall, moderate alignment strength improves consistency and visual quality, whereas excessive supervision may cause artifacts or semantic drift.
\subsection{Instruction Editing}
\begin{figure*}[!h]
\centering
\includegraphics[width=2.0\columnwidth]{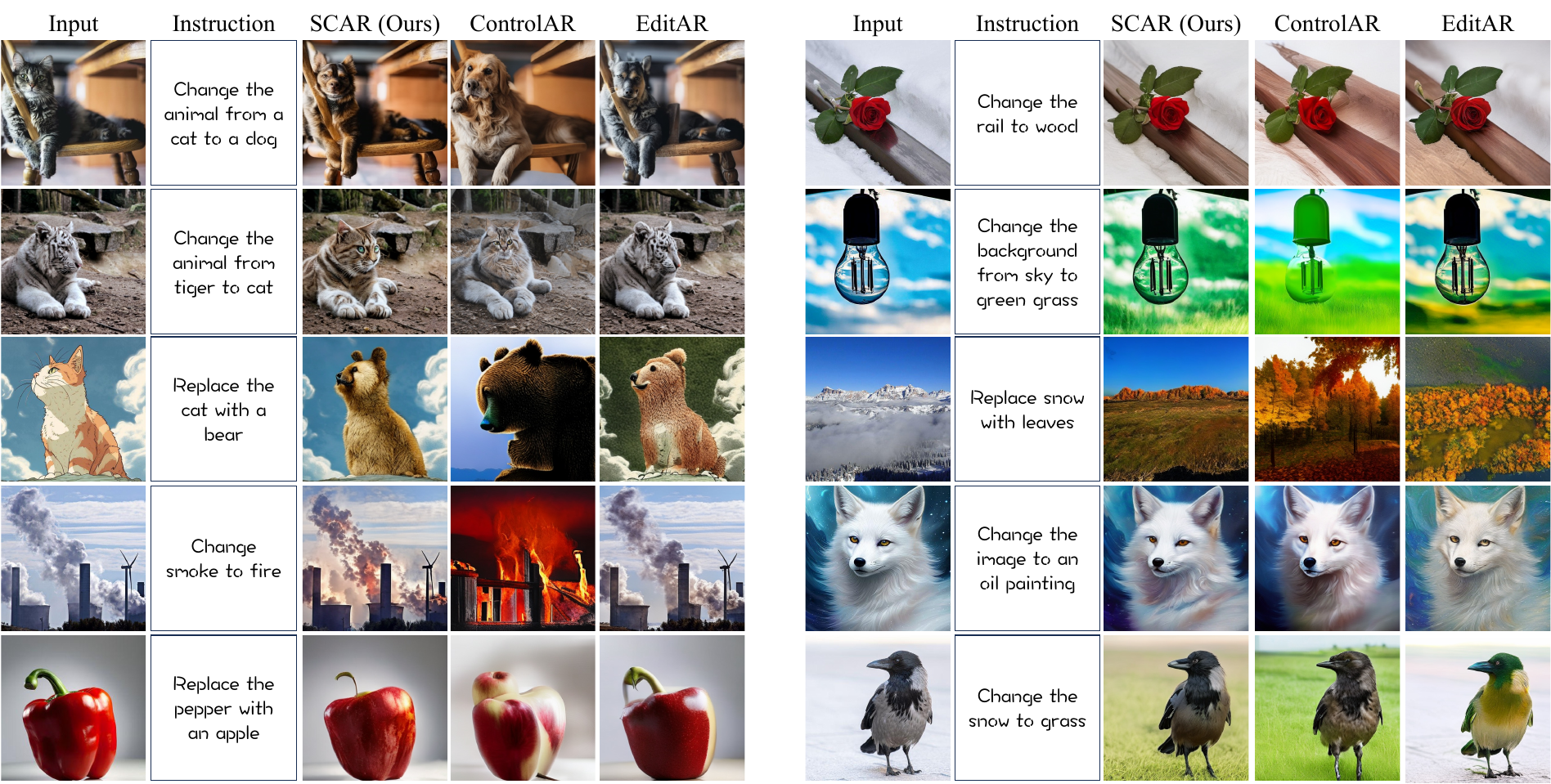}
\caption{
Additional visualizations of instruction editing results. 
SCAR (Ours) produces more faithful and semantically consistent edits than ControlAR~\cite{li2024controlar} and EditAR~\cite{mu2025editar}, with all methods using the same LlamaGen-XL~\cite{llamagen}.
All visualizations are generated at a resolution of 512$\times$512.
}
\vspace{3em}
\label{fig:t2i_edit_app}
\end{figure*}

As shown in~\Cref{fig:t2i_edit_app}, we present additional visualizations for instruction editing, further demonstrating the effectiveness of our proposed SCAR.

\subsection{Controllable Generation}
\begin{figure*}[!h]
\centering
\includegraphics[width=2.0\columnwidth]{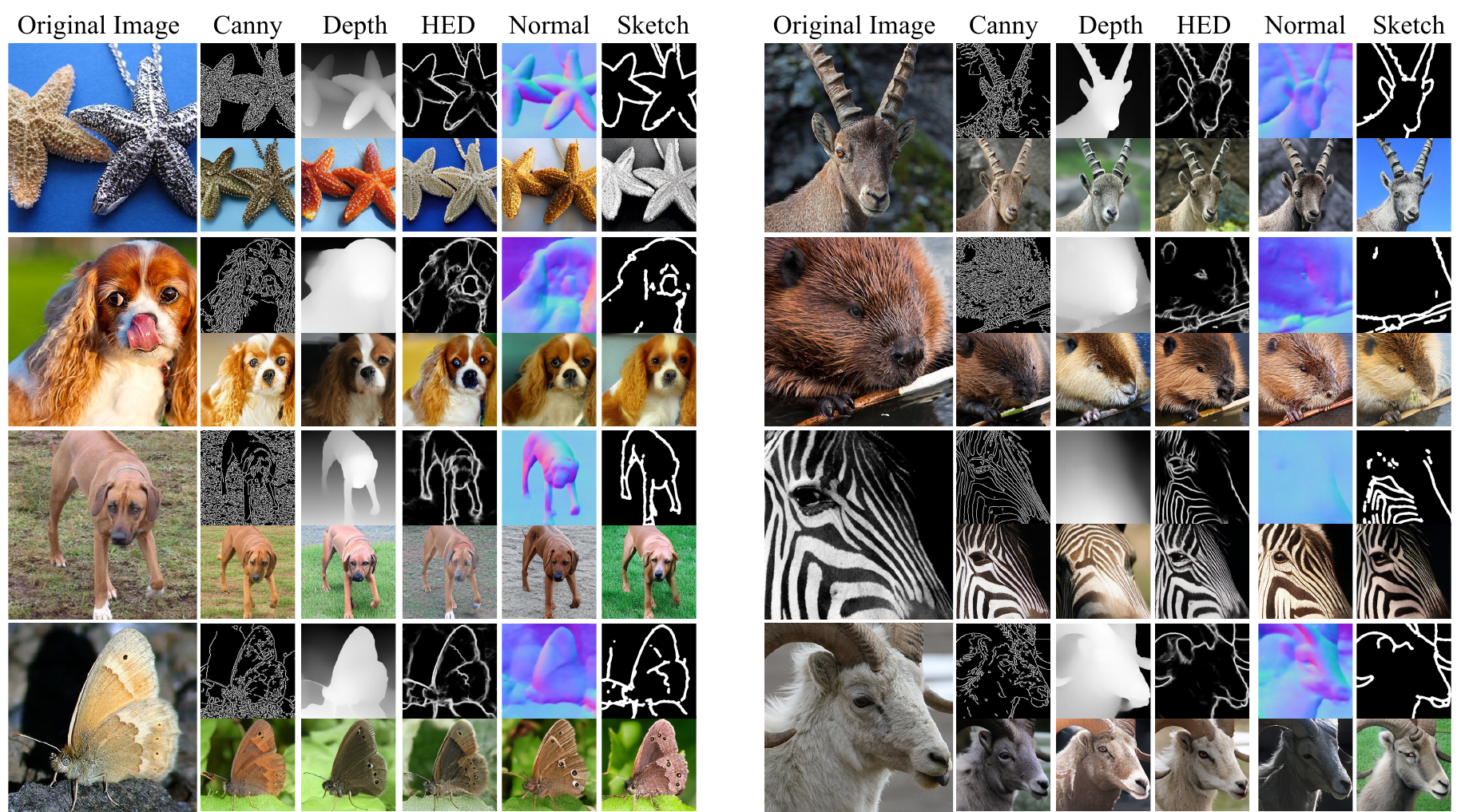}
\caption{
Additional visualizations of C2I Controllable Generation by SCAR-Uni based on LlamaGen-L.
We adopt five different control conditions: Canny, Depth, HED, Normal, and Sketch.
All visualizations are generated at a resolution of 256$\times$256.
}
\label{fig:c2i_control_uni_app}
\end{figure*}

\noindent \textbf{C2I Controllable Generation by SCAR-Uni.}
In C2I Controllable Generation, we focus on the results of SCAR-Uni, a unified method that supports diverse control conditions using the same model. As shown in~\Cref{fig:c2i_control_uni_app}, SCAR-Uni generates high-quality images while maintaining strong controllability across diverse input types.

\begin{figure*}[!h]
\centering
\includegraphics[width=2.0\columnwidth]{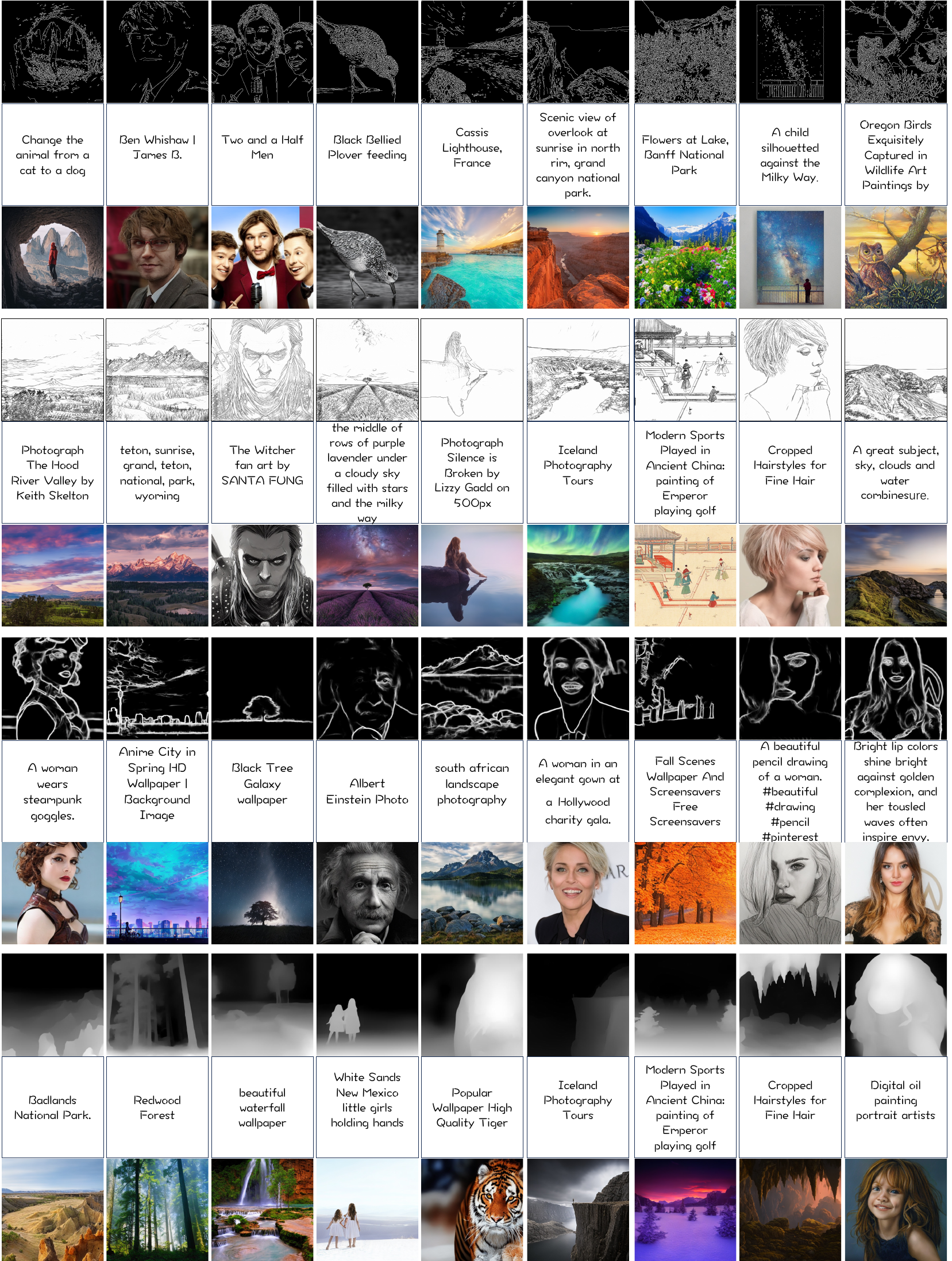}
\caption{
Additional visualizations of T2I Controllable Generation by SCAR based on LlamaGen-XL.
To demonstrate the controllability and generalization ability of our SCAR, we present results under four control conditions: Canny, Depth, HED, and Lineart. 
All images are generated at a resolution of 512×512.
}
\label{fig:t2i_control_app}
\end{figure*}

\noindent \textbf{T2I Controllable Generation.}
\Cref{fig:t2i_control_app} presents additional visualizations for T2I controllable generation. 
The results further demonstrate the strong performance and generalization ability of our proposed SCAR.

\newpage

\clearpage  


\end{document}